\documentclass{aacd}

\usepackage{amsmath}

\usepackage[pdftex]{graphicx}
\usepackage[xindy,toc]{glossaries}
\usepackage{tikz}
\usepackage{tikz-timing}
\usepackage{hyphenat}
\usepackage[utf8]{inputenc}
\usepackage[T1]{fontenc}
\usepackage{textcomp}
\usepackage{prettyref}
\usepackage[hidelinks]{hyperref}
\usepackage{xifthen}
\usepackage{listings}
\usepackage{amsmath}
\usepackage{listings}
\usepackage{hyphenat}
\usetikzlibrary{calc,shapes,fit,positioning,arrows}

\usepackage[EULERGREEK]{sansmath}

\usepackage[binary-units,detect-all]{siunitx}

\usepackage{changes}
\usepackage[colorinlistoftodos,obeyDraft,obeyFinal,final]{todonotes}
\usepackage{cleveref}
\crefname{figure}{fig.}{figs.}
\Crefname{figure}{Fig.}{Figs.}
\crefname{equation}{}{}
\crefname{section}{Section}{Sections} \crefname{subsection}{Section}{Sections} \Crefname{subsection}{Section}{Sections} 
\newcommand {\tdi}[1]{}

\robustify{\gls}
\robustify{\glspl}

\lstdefinestyle{tclstyle}{
  language   = tcl,
  basicstyle = \ttfamily\scriptsize,
	basewidth  = {.5em,0.4em},
	columns    = flexible,
  keepspaces = true,
}
\lstdefinestyle{pythonstyle}{
  language   = python,
  basicstyle = \ttfamily,
	basewidth  = {.5em,0.4em},
	columns    = flexible,
  keepspaces = true,
}

\makeglossaries

\newacronym{adc}{ADC}{Analog to Digital Converter}
\newacronym{dac}{DAC}{Digital to Analog Converter}
\newacronym{ppu}{PPU}{Plasticity Processing Unit}
\newacronym{cpu}{CPU}{Central Processing Unit}
\newacronym{isa}{ISA}{Instruction Set Architecture}
\newacronym{fifo}{FIFO}{First In First Out}
\newacronym{fpga}{FPGA}{Field Programmable Gate Array}
\newacronym{fwhm}{FWHM}{full width at half maximum}
\newacronym{serdes}{SerDes}{Serializer/Deserializer}
\newacronym{simd}{SIMD}{Single Instruction Multiple Data}
\newacronym{sram}{SRAM}{Static Random Access Memory}
\newacronym{stdp}{STDP}{Spike-timing dependent plasticity}
\newacronym{io}{IO}{Input/Output}
\newacronym{dls}{HICANN-DLS}{HICANN-DLS}
\newacronym{lsb}{LSB}{least significant bit}
\newacronym{inl}{INL}{integral nonlinearity}
\newacronym{msb}{MSB}{most significant bit}
\newacronym{gpu}{GPU}{Graphics Processing Unit}
\newacronym{asic}{ASIC}{Application Specific Integrated Circuit}
\newacronym{hicann}{HICANN}{High Input Count Analog Neural Network}
\newacronym{cmos}{CMOS}{Complementary Metal-Oxid-Semiconductor}
\newacronym{lvstl}{LVSTL}{Low-Voltage-Swing Terminated Logic}
\newacronym{nmda}{NMDA}{N-Methyl-D-Aspartat}
\newacronym{psp}{PSP}{post-synaptic potential}
\newacronym{ota}{OTA}{operational transconductance amplifier}
\newacronym{adex}{AdEx}{Adaptive-Exponential Integrate-and-Fire}
\newacronym{hbp}{HBP}{Human Brain Project}
\newacronym{hicanndls}{HICANN-DLS}{High Input Count Analog Neural Network with Digital Learning System}
\newacronym{dlssr}{HICANN-DLS-SR}{High Input Count Analog Neural Network with Digital Learning System as Spikey Replacement}
\newacronym{hagen}{HAGEN}{Heidelberg AnaloG Evolvable Neural network}
\newacronym{pcb}{PCB}{printed circuit board}
\newacronym{anncore}{ANNCORE}{Analog Network Core}
\newacronym{mpw}{MPW}{Multi-Project Wafer}
\newacronym{l1}{Layer1}{Event Link Layer 1}
\newacronym{l2}{Layer2}{Event Link Layer 2}
\newacronym{spl1}{spL1}{Synchronous Parallel Event Link Layer 1}
\newacronym{lvds}{LVDS}{Low Voltage Differential Signalling}
\newacronym{crc}{CRC}{Cyclic Redundancy Check}
\newacronym{hicannxfull}{HICANN-X}{HICANN-DLS-SR-HX}
\newacronym{hicannx}{HICANN-X}{HICANN-X}
\newacronym{mos}{MOS}{Metal-Oxid Semiconductor}
\newacronym{fet}{FET}{Field-Effect Transistor}
\newacronym{mosfet}{MOSFET}{Metal-Oxid Semiconductor Field-Effect Transistor}
\newacronym{iaf}{I\&F}{Integrate-and-Fire}
\newacronym{dtc}{DTC}{digital-to-time converter}
\newacronym{cadc}{CADC}{Correlation-readout ADC}
\newacronym{hpc}{HPC}{high-performance computing}
\newacronym{bss2}{BSS-2}{BrainScales-2}
\newacronym{bss1}{BSS-1}{BrainScales-1}
\newacronym{bss}{BSS}{BrainScales}
\newacronym{cdnn}{CDNN}{Convolutional Deep Neural Network}
\newacronym{soc}{SOC}{Sytem On Chip}

\newacronym{padi}{PADI}{parallel driver interface}

\newacronym{dut}{DUT}{design under testing}
\newacronym{ocp}{OCP}{Open Core Protocol}
\newacronym{stp}{STP}{Short Time Plasticity}
\newacronym{sta}{STA}{static timing analysis}
\newacronym{gals}{GALS}{globally asynchronous locally synchronous}
\newacronym{dpi}{DPI}{SystemVerilog direct programming interface}
\newacronym{pll}{PLL}{phase-locked loop}

\newacronym{psc}{PSC}{post-synaptic current}
\newacronym{rstdp}{R-STDP}{reward-modulated spike-timing-dependent plasticity}
\newacronym{lif}{LIF}{leaky integrate-and-fire}

\newacronym{mc}{MC}{Monte Carlo}

\makeatletter
\renewcommand*{\@seccntformat}[1]{   \csname the#1\endcsname.\quad
}
\makeatother

\begin{document}
\title{Accelerated Analog Neuromorphic Computing}
\author{Johannes~Schemmel, Sebastian~Billaudelle, Phillip~Dauer, Johannes Weis\\
Heidelberg University, Heidelberg, Germany
\\\{schemmel\}@kip.uni-heidelberg.de}
\maketitle
\thispagestyle{empty}
\todo{
6 to 20 pages
}
\begin{abstract}
This paper presents the concepts behind the \gls{bss} accelerated analog neuromorphic computing architecture.
It describes the second-generation \gls{bss2} version and its most recent in-silico realization, the HICANN-X \gls{asic}, as it has been developed as part of the neuromorphic computing activities within the European \gls{hbp}. 
While the first generation is implemented in an \SI{180}{\nano \meter} process, the second generation uses \SI{65}{\nano \meter} technology.
This allows the integration of a digital plasticity processing unit, a highly-parallel micro processor specially built for the computational needs of learning in an accelerated analog neuromorphic systems.

The presented architecture is based upon a continuous-time, analog, physical model implementation of neurons and synapses, resembling an analog neuromorphic accelerator attached to build-in digital compute cores.
While the analog part emulates the spike-based dynamics of the neural network in continuous-time, the latter simulates biological processes happening on a slower time-scale, like structural and parameter changes. Compared to biological time-scales, the emulation is highly accelerated, i.e. all time-constants are several orders of magnitude smaller than in biology.
Programmable ion channel emulation and inter-compartmental conductances allow the modeling of non-linear dendrites, back-propagating action-potentials as well as NMDA and Calcium plateau potentials.
To extend the usability of the analog accelerator, it also supports vector-matrix multiplication.
Thereby, \gls{bss2} supports inference of deep convolutional networks as well as local-learning with complex ensembles of spiking neurons within the same substrate.
A prerequisite to successful training is the calibratability of the underlying analog circuits across the full range of process variations.
For this purpose a custom software toolbox has been developed, that facilitates complex calibrated Monte-Carlo simulations.
\end{abstract}

\section{Introduction\label{sec:intro}}
The basic concept of the BrainScaleS systems is the emulation of biologically-inspired neural networks with physical models\cite{schemmel2010iscas}. 
It differs from comparable neuromorphic approaches based on continuous-time analog circuits\cite{indiveri2011neuromorphic, benjamin2014neurogrid, douglas95neuromorphic} in many aspects, like the high acceleration factor\cite{schemmel_ijcnn06,schemmel_iscas07}, usage of wafer-scale integration\cite{zoschke2017full}, calibratability towards biologically-sound neuron parameters\cite{millner10,pfeil2013six}, a software-interface based on the simulator-agnostic description language PyNN\cite{davison08pynn,bruederle_biolcybern2010}, support for non-linear dendrites and structured neurons\cite{schemmel2017nmda} as well as on-chip support for complex plasticity rules based on a combination of analog measurements  internal analog-to-digital conversion and build-in microprocessors.

The first generation, BrainScaleS~1, has been completed\cite{thakur2018large} and is used mostly for research of connectivity aspects of large accelerated analog neural networks and the further development of wafer-scale integration technology. 
\begin{figure}
    \center{      \includegraphics[width=0.9\linewidth,page=25,viewport=0 0 17.5cm 9cm, clip]{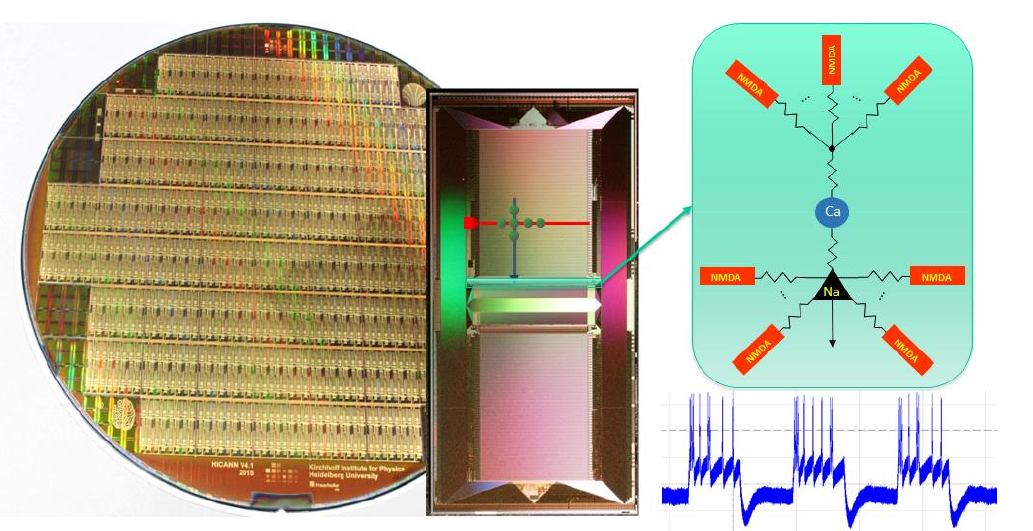}
      \caption{Basic elements of the BrainScaleS architecture: wafer, \gls{bss1} \gls{asic}, \gls{bss2} neuron and exemplary membrane voltage trace.
      \label{fig:bss} }}
\end{figure}
The main short-coming of the BrainScaleS~1 system is the rather inflexible implementation of long-term plasticity based solely on \gls{stdp}, which has been taken over from its predecessor\cite{schemmel_iscas07}. 
Already at the very beginning of the BrainScaleS project this was considered a conceptual weakness and an upgrade path was devised to implement the more flexible hybrid plasticity\cite{friedmann2016hybridlearning} scheme in future revisions. 
Due to the process technology used within BrainScaleS~1, \SI{180}{nm}, it was not feasible to integrate the necessary standard cell logic without sacrificing too much area to digital circuits in relation to the analog neurons and synapses. 
Therefore, the decision was made to develop a second BrainScaleS generation, BrainScaleS~2, which is based from the beginning on a smaller process technology, namely \SI{65}{nm}. 
Fig.~\ref{fig:bss} shows the main elements of the \gls{bss} architecture.
At the very left, a \gls{bss1} wafer containing approx. 500 interconnected \glspl{asic} is shown.
To its right, a \gls{bss} chip illustrates the characteristic layout of \gls{bss} neuromorphic chips: a central neuron area surrounded by two large synapse blocks.
The sketched overlay shows the rectangular orientation of input (pre-synaptic) and output (post-synaptic) signals: the input is routed horizontally through the synapse array, while the output of the synapses connects them vertically to the neurons in the center.
Next to it the graphical representation of an emulated structured neuron is shown above a measured voltage-trace from the membrane capacitor of a neuron.

One major improvement is the inclusion of a digital plasticity processor in the \gls{bss2} \gls{asic}\cite{friedmann2016hybridlearning}. 
This specialized highly-parallel \gls{simd} microprocessor adds an additional layer of modeling capabilities, covering all aspects of structural and parameter changes during network operation. 
By including the necessary logic directly within the analog network core, a communication bottleneck to the host system is avoided. 
This allows to scale-up all novel plasticity features for wafer-scale integration within the BrainScaleS~2 system. 
In the finale multi-wafer version of the BrainScaleS~2 system, which is planned to be capable of extending experiments across several hundreds of wafers, the distributed local compute capability will be even more essential. 
It will not only perform all levels of plasticity calculations, but also the initialization and calibration of the numerous analog mixed-signal circuits within the \gls{asic}.
\begin{figure}
    \center{      \includegraphics[width=0.8\linewidth,page=23,viewport=0 0 25.5cm 15cm, clip]{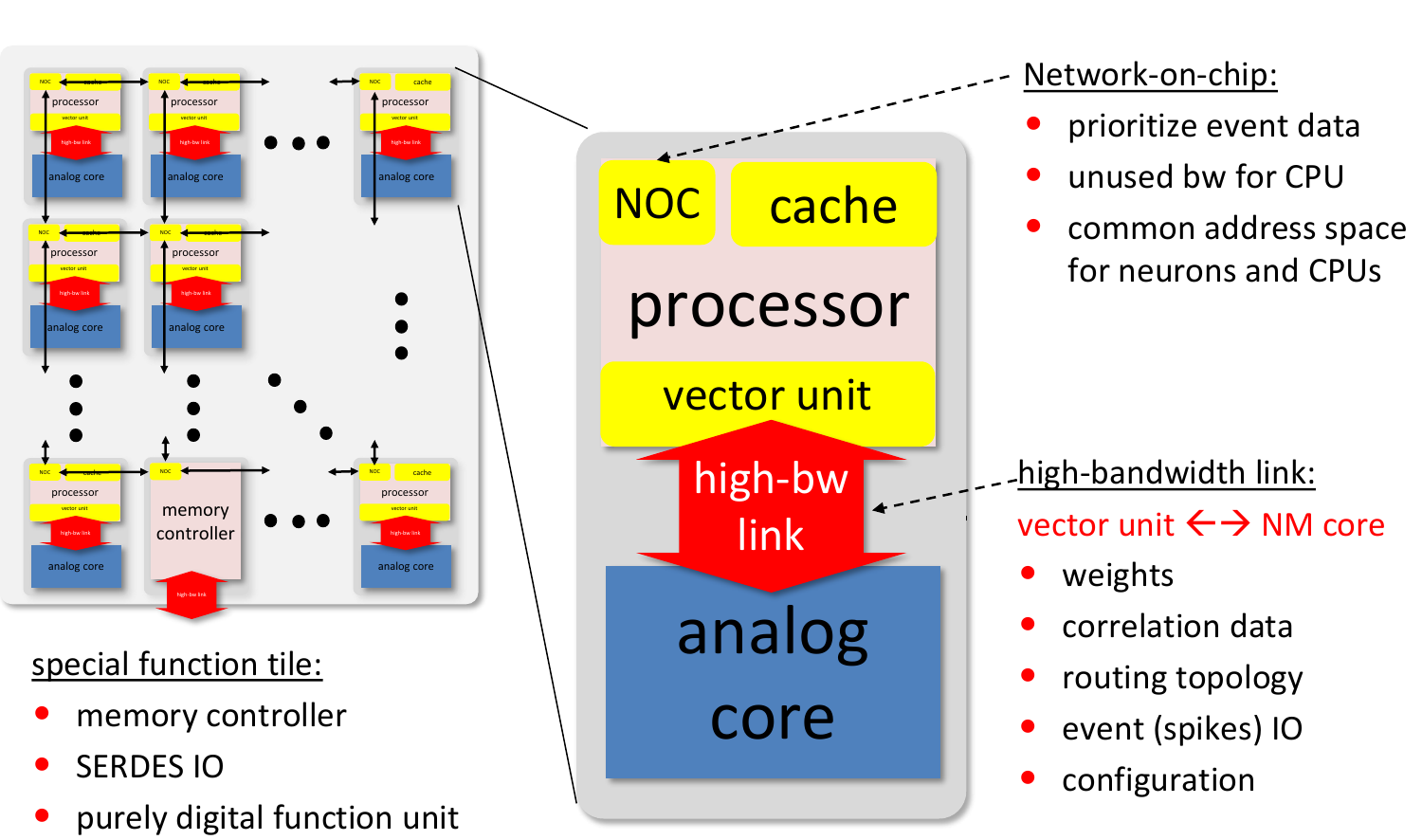}
      \caption{A neuromorphic \gls{soc} consisting of a multitude of digital \gls{cpu} cores with special vector units attached to analog Neuromorphic accelerators. 
      \label{fig:multicore} }}
\end{figure}
The role of the analog neural network block changes by the transition from \gls{bss1} to \gls{bss2}.
The analog part becomes an attachment to the \gls{cpu} cores, similar to a complex accelerator. 
Fig.~\ref{fig:multicore} illustrates this architecture.

The remainder of this publication is organized as follows: Section~\ref{sec:bss} gives an overview of the \gls{bss2} architecture.
Section~\ref{sec:dlsx} presents the current prototype, the single-chip variant of \gls{bss2}, called \gls{hicannx}. 
Section~\ref{sec:teststand} shows some examples of the complex calibrated Monte-Carlo simulations used to verify that the analog neurons circuits are always capable of correctly emulating their biological counterparts, i.e. their calibratability under all process and device variations.
The paper closes with a conclusion in Section~\ref{sec:conclusion}.

\section{Overview of the \gls{bss} neuromorphic architecture\label{sec:bss}}
As shown in Fig.~\ref{fig:multicore}, the \gls{bss} architecture is based on the close interaction of digital and analog circuit blocks. Because of their primary intended function, the digital processor cores are called \glspl{ppu}. 
As the main neuromorphic component, the analog core contains synapse and neuron circuits \cite{aamir16dlsneuron,aamir2018accelerated}, analog parameter memories, \gls{ppu} interfaces as well as all event related interface components. 

The \gls{ppu} is an embedded microprocessor core with a highly parallel \gls{simd} unit optimized for the calculation of plasticity rules in conjunction with the analog core\cite{friedmannschemmel2016}.
In the current incarnation of the \gls{bss} architecture, \gls{bss2}, two \glspl{ppu} share an analog core.
This allows the most efficient arrangement of the neuron circuits in the center of the analog core.
\begin{figure}
    \center{      \includegraphics[width=0.8\linewidth,page=4,viewport=0 0 21.3cm 15.3cm, clip]{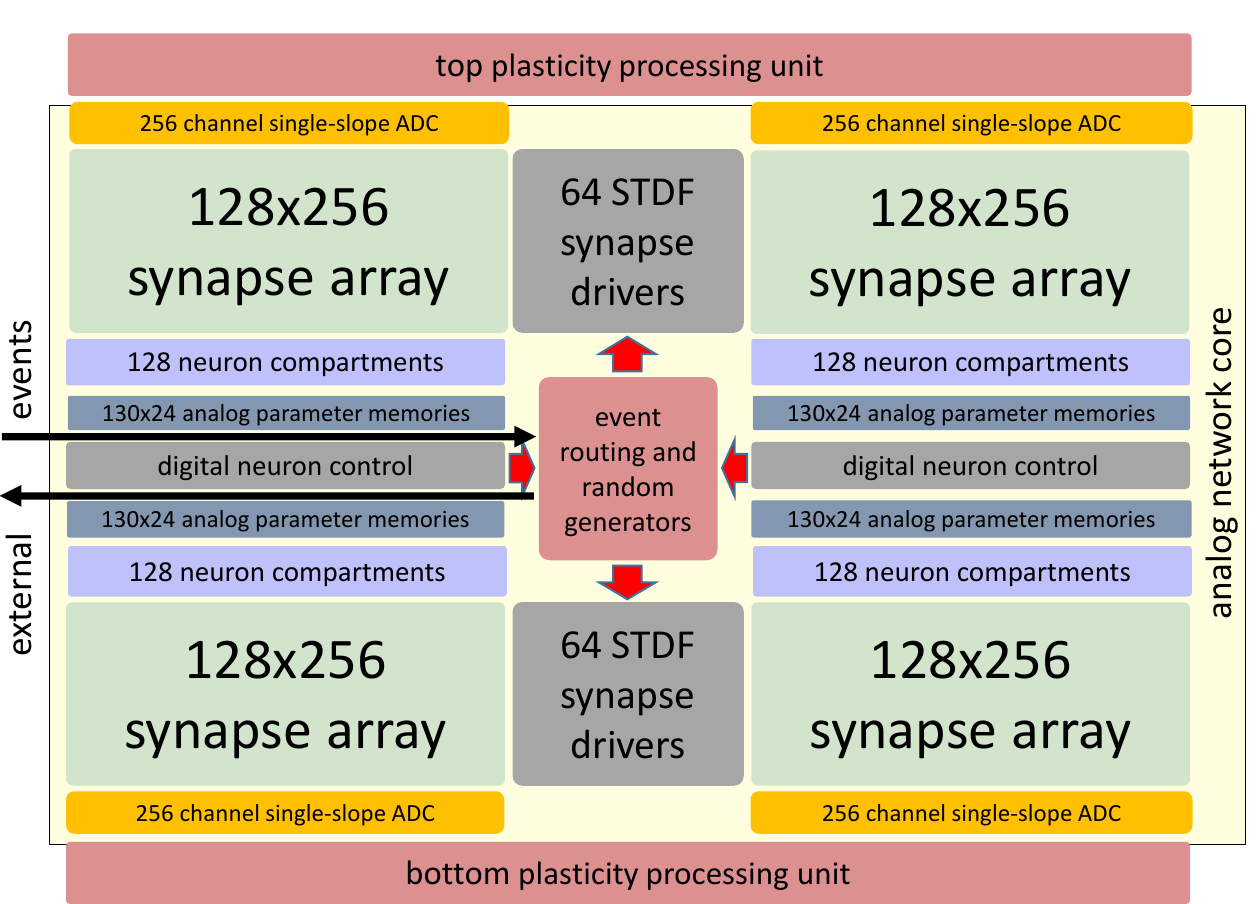}
      \caption{Block diagram of the \gls{anncore}.
      \label{fig:anncorebd} }}
\end{figure}
Fig.~\ref{fig:anncorebd} depicts the individual function blocks located within the \gls{anncore}:
\begin{description}
\item[synapse arrays]\hfill \\
The total number of synapses are split up in four equally sized blocks to keep the vertical and horizontal lines traversing the sub-arrays as short as possible, thereby reducing their parasitic capacitances (see \cite{friedmannschemmel2016,aamir2018accelerated}). Each synapse array resembles a block of static memory, with 16 memory cells located in each synapse, organized in two words of eight bits each. A synapse array also contains the sense amplifiers, precharge and write control circuits as well as word-line decoders and buffers.
Thereby it can be connected directly to the digital, standard cell based parts of the chip.
Two \glspl{ppu} connect to the static memory interfaces of the two adjacent synapse arrays, using a fully parallel connection to the 8$\times$256 data lines. 
\pagebreak
\item[neuron compartment circuits]\hfill \\
Four rows of neuron compartment circuits are located at the edges of the synapse blocks.
Each pair of dendritic input lines of a neuron compartment is connected to a column of 256 synapses.
The neuron compartment implements the \gls{adex} neuron model. They can be connected to form larger neurons, emulating either point or structured neurons.
See \cite{schemmel2017nmda} for more details about the multi-compartment capabilities.
\item[analog parameter memories]\hfill \\
Adjacent to each row of neuron compartments is a row of analog parameter storages. These capacitive memories \cite{hock13analogmemory} store 24 analog values per neuron and an additional 48 global parameters
They are auto-refreshed from values stored digitally inside the memory block.
\item[digital neuron control]\hfill \\
Two neuron rows share a digital neuron control block which synchronizes neural events to the digital system clock of \SI{125}{\mega\hertz} and serializes them onto digital output buses.
\item[synapse drivers with short term plasticity]\hfill \\
The pre-synaptic events are fed into the array via the synapse drivers. Besides timing control and buffering they contain short-term plasticity circuits emulating a simplified Tsodys-Markram model \cite{tsodyks97neural,schemmel_iscas07}.
The synapse drivers can handle single- or multi-valued input signals, depending on the current operation mode of the synapse row, which may be either rate or spike based.
\item[random event generators]\hfill \\
The random generators produce random background events fed directly into the synapse array via the synapse drivers, strongly reducing the external bandwidth usage when stochastic models \cite{pfeil2014effect,jordan2019deterministic} are used.
\item[correlation \glspl{adc}]\hfill \\
The top and bottom edges of the \gls{anncore} are lined by the \gls{simd} units of the top and bottom \glspl{ppu}. A column-parallel \glspl{adc} converts the analog data from the synapse arrays as well as selected analog signals from the neurons into the digital representations needed by the \glspl{ppu} . 
\end{description} 
\begin{figure}
    \center{      \includegraphics[width=1.0\linewidth,page=5,viewport=0 0 22.8cm 18cm, clip]{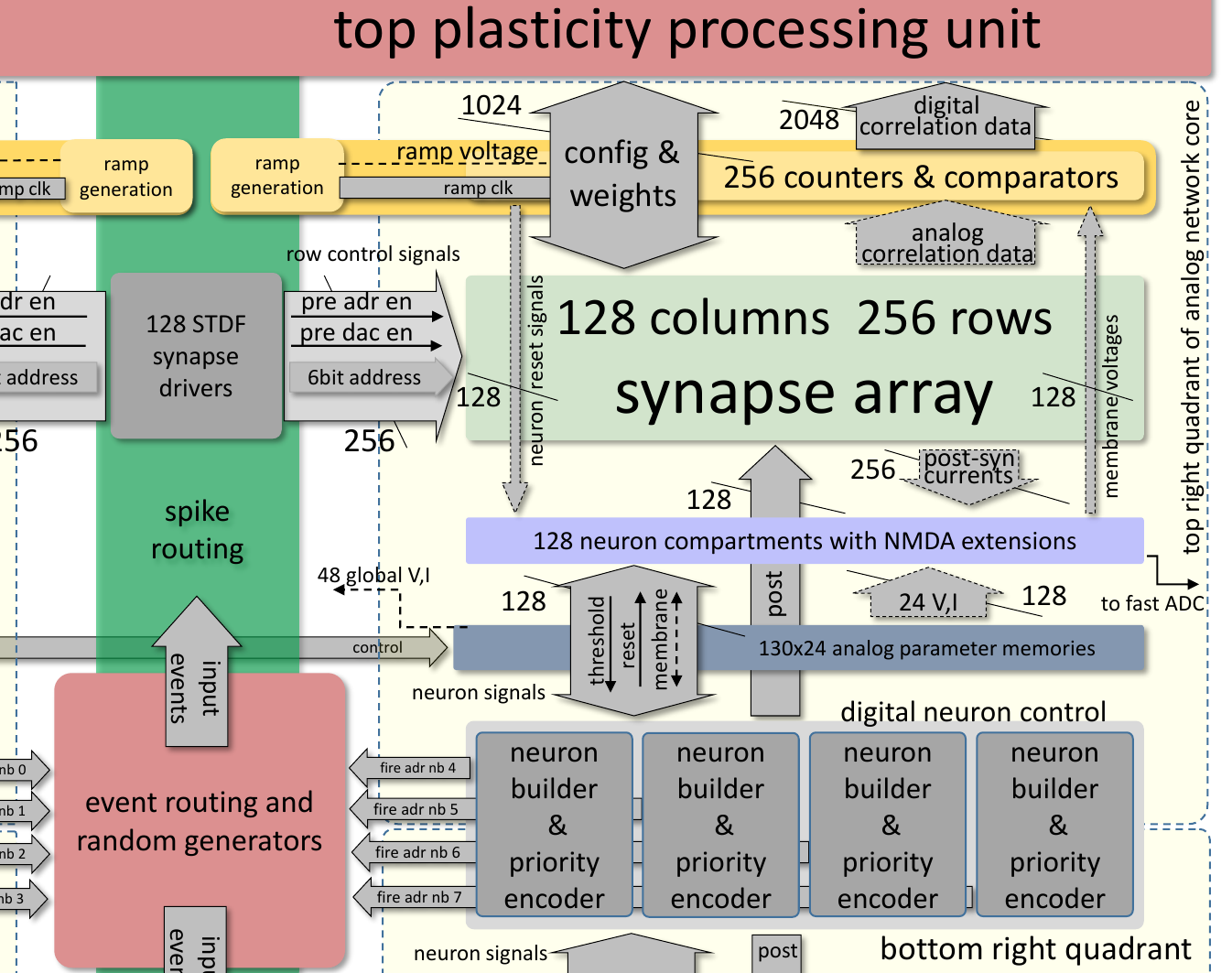}
      \caption{Detailed block diagram of the \gls{anncore}'s upper right quadrant.
      \label{fig:anncoretrbd} }}
\end{figure}
Fig.~\ref{fig:anncoretrbd} shows a zoom-in into the upper right quadrant of the \gls{anncore}. For compatibility with \gls{bss1}, the synapse drivers and digital neuron control circuits are arranged in a similar substructure as they have been previously: one synapse driver controls two rows of synapses in both adjacent blocks and the digital neuron control is split in eight blocks controlling 64 neuron compartments each.
Four blocks are located in the left and four in the right half of \gls{anncore}. 
Each block contains the so-called neuron builder logic, which allows to interconnect analog membrane and digital spike output signals from neuron compartments being either vertically or horizontally adjacent to each other. 
To serialize the up-to 64 spike outputs each digital neuron control block contains priority encoder circuits that arbitrate the access to the output bus. It also contains a 8 $\times$ 64 neuron source address memory\cite{kiene2017masterthesis}.

The pre-synaptic input for the synapse drivers of one chip half comes from a set of local event input buses driven by the central event router. The event router within the \gls{anncore} mixes global, local and random event sources.
\begin{figure} 
    \center{      \includegraphics[width=0.8\linewidth,page=13,viewport=0 0 21.6cm 160mm, clip]{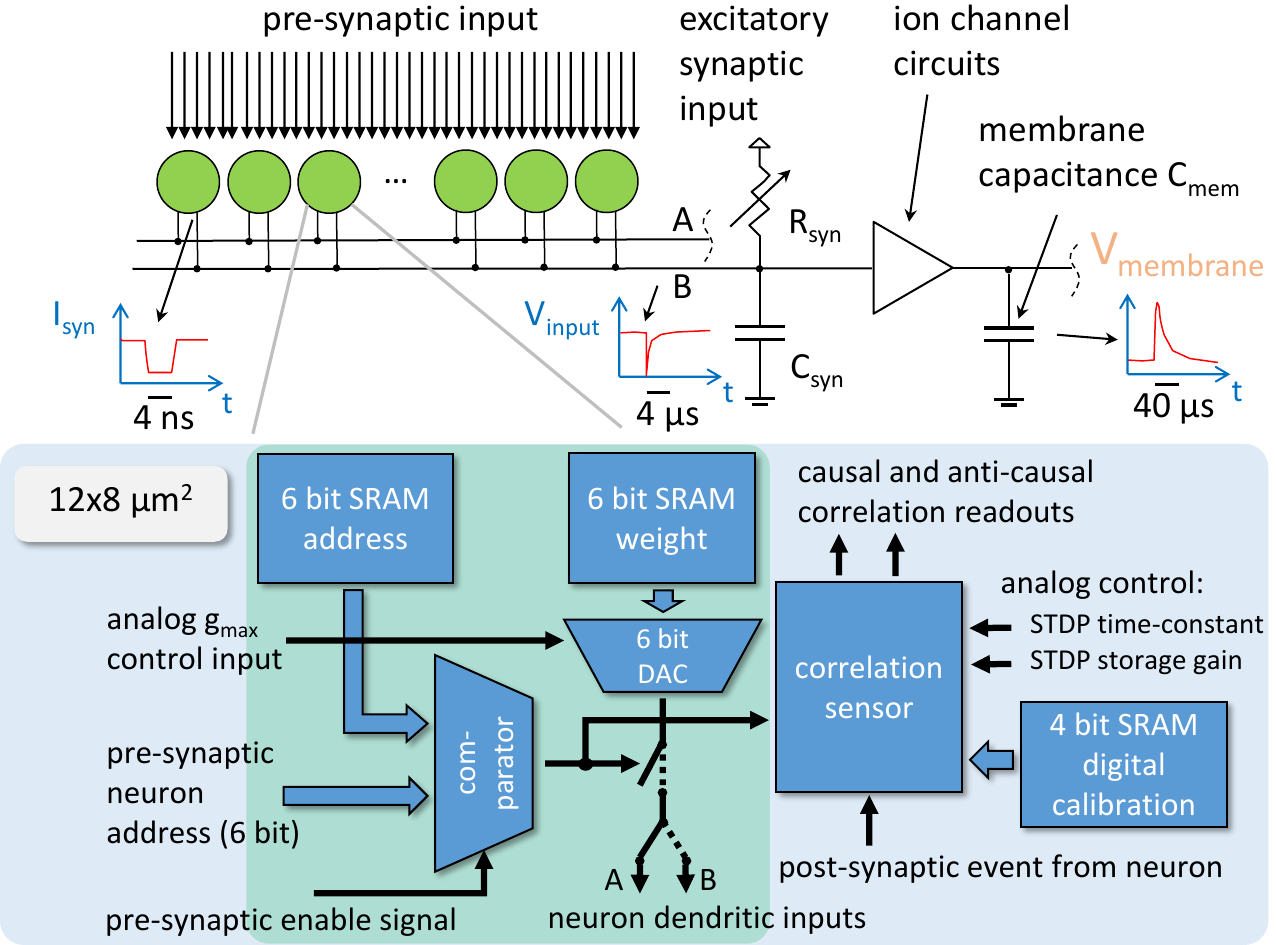}
      \caption{Top: Operating principle and basic timing relationships of an accelerated BrainScaleS spiking neuron. Bottom: Block-diagram of a synapse.\label{fig:opneuron} }}
\end{figure}
In Fig.~\ref{fig:anncoretrbd} the synapses are arranged in a two-dimensional array between the \gls{ppu} and the neuron compartment circuits.
Pre-synaptic input enters the synapse array at the left edge.
For each row, a set of signal buffers transmit the pre-synaptic pulses to all synapses in the row.
The post-synaptic side of the synapses, i.e. the equivalent of the dendritic membrane of the target neuron, is formed by wires running vertically through each column of synapses.
At each intersection between pre- and post-synaptic wires, a synapse is located.
To avoid that all neuron compartments share the same set of pre-synaptic inputs, each pre-synaptic input line transmits - in a time-multiplexed fashion - the pre-synaptic signals of up to 64 different pre-synaptic neurons.
Each synapse stores a pre-synaptic address that determines the pre-synaptic neuron it responds to.

Fig.~\ref{fig:opneuron} illustrates the basic operation of the BrainScaleS accelerated analog neuron and its associated synapses.
Due to space limitations the dendritic column is rotated by \SI{90}{\degree} in the figure. The bottom half of the figure shows a block diagram of the synapse circuit.
The main functional blocks are the address comparator, the \gls{dac} and the correlation sensor.
Each of these circuits has its associated memory block.
The address comparator receives a \SI{6}{bit} address and a pre-synaptic enable signal from the periphery of the synapse array as well as a locally stored \SI{6}{\bit} neuron number.
If the address matches the programmed neuron number, the comparator circuit generates a pre-synaptic enable signal local to the synapse ({\em pre}), which is subsequently used in the \gls{dac} and correlation sensor circuits. 
Each time the \gls{dac} circuit receives a {\em pre} signal, it generates a current pulse.
The height of this pulse is proportional to the stored weight, while the pulse width is typically \SI{4}{ns}.
This matches the maximum pre-synaptic input rate of the whole synapse row which is limited to \SI{125}{MHz}.
The remaining \SI{4}{ns} are necessary to change the pre-synaptic address.
The current pulse can be shortened below the \SI{4}{ns} maximum pulse length to emulate short-term synaptic plasticity \cite{schemmel_iscas07,billaudelle2017stp}.

Each neuron compartment has two inputs, labeled A and B in Fig.~\ref{fig:opneuron}. 
Usually, the neuron compartment uses A as excitatory and B as inhibitory input.
Each row of synapses is statically switched to either input A or B, meaning that all pre-synaptic neurons connected to this row act either as excitatory or inhibitory inputs to their target neurons.
Due to the address width of \SI{6}{bit} the maximum number of different pre-synaptic neurons is 64\cite{billaudelle2019structural}. 
The output currents of all synapses discharge the synaptic input capacitance $C_\textrm{syn}$, which is realized predominantly by the shielding capacitance of the long synaptic input wires. An adjustable \gls{mos} resistor,$R_\textrm{syn}$, restores the charge. 
Due to the short time-constant of the synaptic input pulse compared to the time constant of the synaptic input line $\tau_\textrm{input}=C_\textrm{syn}R_\textrm{syn}$, which is three orders of magnitude longer, the voltage trace $V_\textrm{input}(t)$ is a single exponential.

The ion-channel circuits in BrainsScaleS should implement the full \gls{adex} neuron model, as it is the case in the \gls{bss1} system.\cite{brette_05,millner2012develop,aamir2018accelerated}.
In \gls{bss2} some terms are still under development at the time of this writing.
The minimum configuration available in all prototype versions of \gls{bss2} is a set of two current-based inputs, one for inhibitory synaptic input, connected to input A in Fig.~\ref{fig:opneuron} and one for excitatory (input B), in combination with a leak circuit and spike and reset generation\cite{aamir16dlsneuron}.
Therefore the membrane voltage is given by the standard \gls{iaf} neuron model\cite{jolivet2004generalized}. 
Typically, the membrane time constant set by the leakage term is another order of magnitude above the time-constant of the synaptic input.
These temporal relationships are visualized in the small timing diagram inserts in Fig.\ref{fig:opneuron}.

The remaining functional block of the synapse shown in Fig.~\ref{fig:opneuron} is the correlation sensor.
Its task is the measurement of the time difference between pre- and post-synaptic spikes. To determine the time of the pre-synaptic spike it is connected to the {\em pre} signal.
The post-synaptic spike-time is determined by a dedicated signaling line running from each neuron compartment vertically through the synapse array to connect to all synapses projecting to inputs A or B of the compartment\cite{friedmannschemmel2016}.

\section{The \gls{hicannx} chip\label{sec:dlsx}}
Although the target of the \gls{bss} architecture is wafer-scale integration, which offers a cost-effective possibility to build brain-size spiking neural network models, smaller solutions based upon single \glspl{asic} are needed to develop and debug the final design.
They also shorten the time to first experiments, of which a significant proportion does need only hundreds up to a few thousand neurons and therefore does not necessarily rely on wafer-scale integration.
Depending of the complexity of the neuron model they utilize, a few tens of interconnected \gls{bss} \glspl{asic} might be sufficient.
To support these goals, an intermediate version of the second-generation \gls{bss} technology has been developed: suited for single- or multi-chip operation, but simultaneously prepared for later wafer-scale integration.
This section will introduce said single-chip version of \gls{bss2}, called \gls{hicannx}, in more detail.
\begin{figure}
    \center{      \includegraphics[width=0.8\linewidth,page=9,viewport=0 0 23cm 15.2cm, clip]{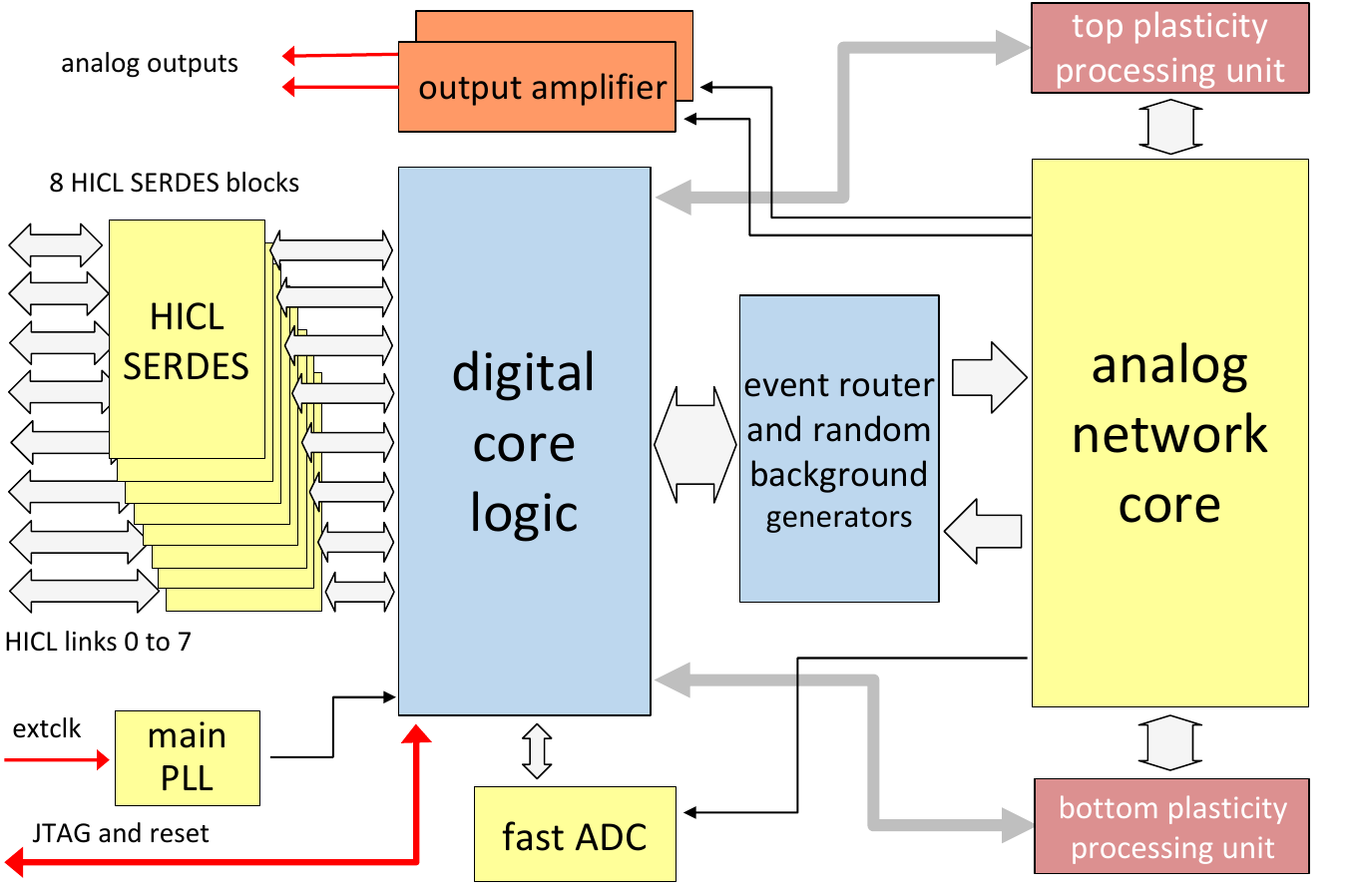}
      \caption{Block diagram of the \gls{hicannx} \gls{asic}.
      \label{fig:dlsxbd} }}
\end{figure}
\begin{figure}
    \center{      \includegraphics[width=1.0\linewidth,page=24,viewport=0 0 25.5cm 13.0cm, clip]{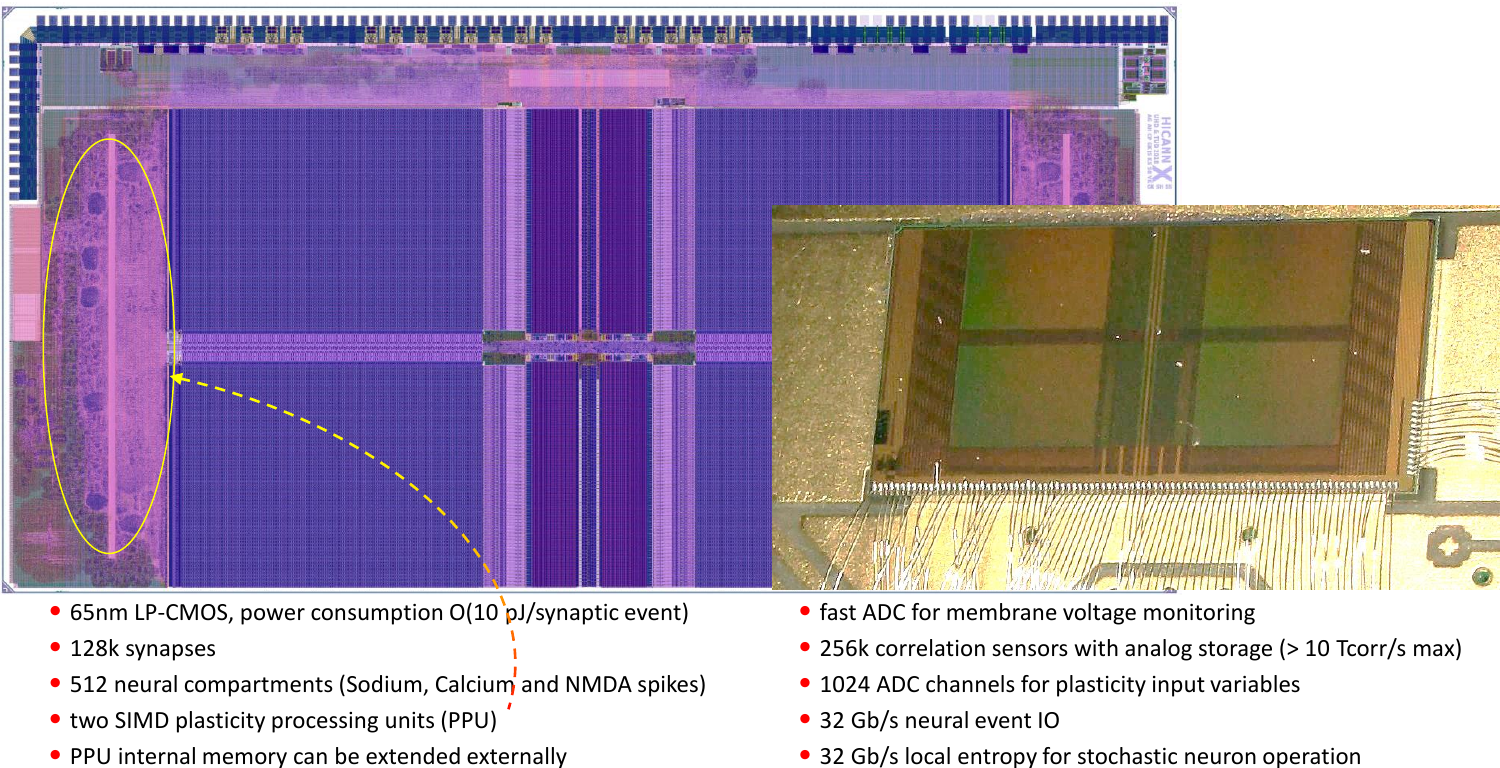}
      \caption{Top: Layout drawing and chip photograph of the \gls{hicannx} \gls{asic}. Bottom: key features of \gls{hicannx}.
      \label{fig:dlsxld} }}
\end{figure}
Fig.~\ref{fig:dlsxbd} shows a block diagram of \gls{hicannx}. 
In total, the \gls{hicannx} chip uses 16 differential \gls{lvds} lines for the host communication.
A single chip has the same bandwidth as the full \gls{bss1} reticle build from eight individual chips.
Using this link arrangement, the \gls{hicannx} chip can be directly connected to one communication module of the BrainScaleS system, providing an easy upgrade path\cite{thanasoulis2012dedicated}.
The layout and photograph of the chip are shown in Fig.~\ref{fig:dlsxld}. 
\subsection{Event-routing within \gls{hicannx} \label{sec:eventdlsx}}
\gls{hicannx} uses the same two-level communication infrastructure as the first \gls{bss} generation\cite{schemmel_ijcnn2008}: a real-time address-event layer without handshake, called \gls{l1}, and a second layer using time-stamped event packets\footnote{The \gls{l1} data format codes a neural event as a parallel bit-field containing the neuron address and a valid bit. It is real-time data with a temporal resolution of the system clock, which is \SI{250}{\mega \hertz} in \gls{hicannx}.}. 
\begin{figure}
    \center{      \includegraphics[width=.6\linewidth,page=22,viewport=0 0 215mm 165mm, clip]{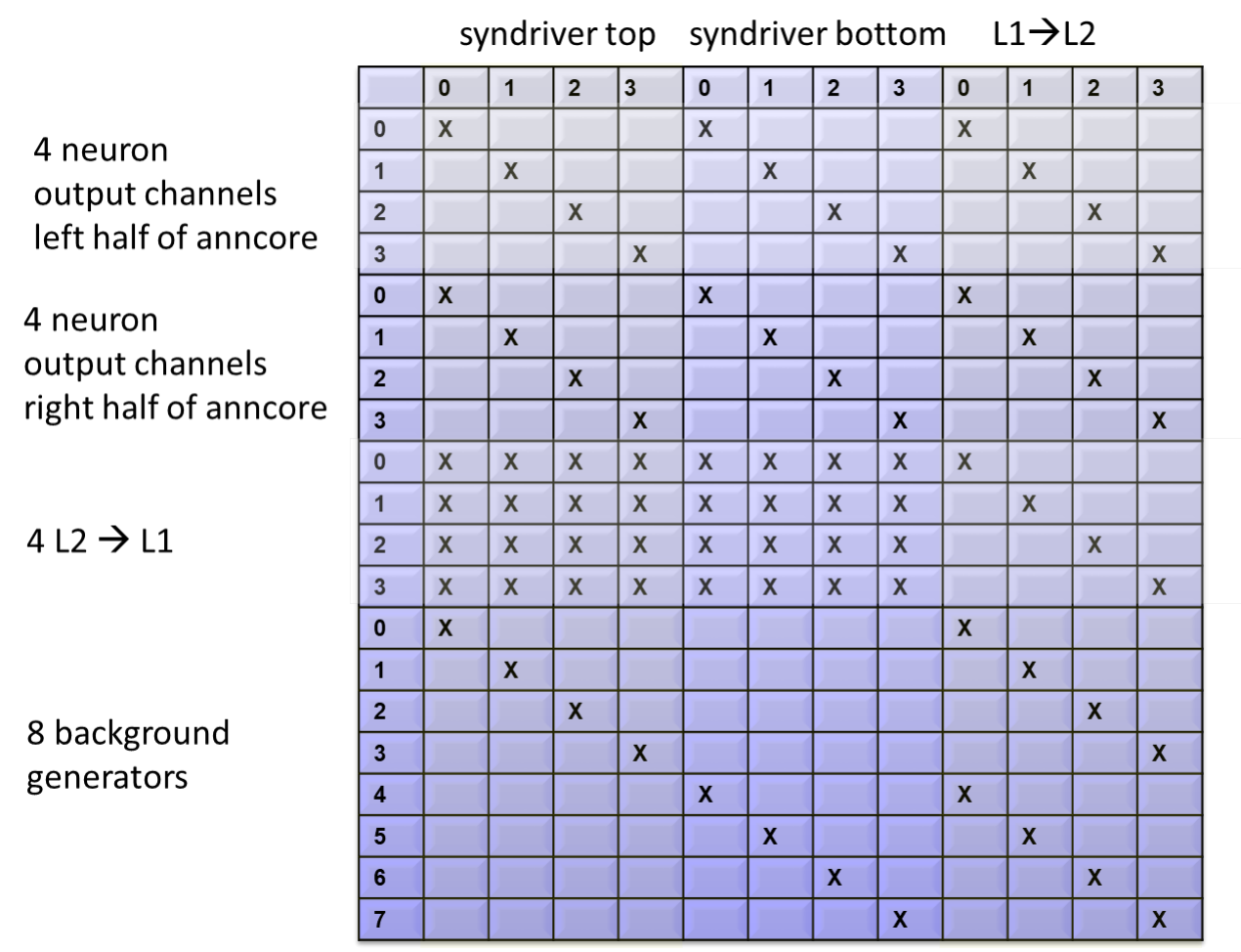}
      \caption{Conceptual view of the internal digital event routing matrix of the \gls{hicannx} chip. All 20 sources a shown vertically on the left, while the 12 output channels are listed at the top. At each position marked with a cross a programmable routing element is located.
    }\label{fig:dlsxspl1} }
\end{figure}
Fig.~\ref{fig:dlsxspl1} shows the implementation of the central \gls{l1} digital event routing network. There are two main sources and sinks for event data: the analog network core, which has eight input event and eight output event buses, as well as the \gls{l2}$\rightarrow$\gls{l1} converter, which provides four links in each direction.
With the exception of the analog core input buses, each link can handle one event per clock cycle of \SI{4}{\nano \second}. The \gls{anncore} input buses are limited to one event every two cycles.

All eight \gls{hicann} compatible links are used for \gls{l2} based event transport. An event is encoded as a combination of neuron address and time stamp. The conversion between time-stamped \gls{l2} data and real-time \gls{l1} data is preformed inside the \gls{l2}$\rightarrow$\gls{l1} converter loacted in the digital core logic. 
It uses a globally synchronized system time counter for this purpose.
The routing of all \gls{l1} events is done within the router matrix. Inside this module are several columns of buffered n-to-1 event merger stages allowing to combine the data of a set of inputs into one \gls{l1} output channel.
All eight physical links of the chip can be simultaneously used for neuron event data (\gls{l2}), slow control and \gls{ppu} global memory accesses. The number of active links might be statically programmed to be any number between one and the maximum of eight. This is useful if several chips should be connected to a single host with a limited number of available links. 
All events transferred via \gls{l2} are protected against undetected bit-errors by \gls{crc} fields.
\subsection{Analog Inference: Rate-based Extension of \gls{hicannx}\label{sec:hagen}}
One of the first neuromorphic systems build in Heidelberg was \gls{hagen}, a fast analog Perceptron-based network chip optimized for hardware-in-the-loop training\cite{schemmel_kluwer04hagen}.
Caused by parallel activities withing the Heidelberg Electronic Vision(s) research group\cite{langeheine_gecco04} it was mainly trained by evolutionary algorithms\cite{hohmann_ijcnn04}, explaining the acronym.
Nevertheless, it was perfectly usable for other hardware-in-the-loop based algorithms, similar to the deep-learning results that have been more recently achieved by other neural network chips used in a Perceptron-like fashion\cite{nurse2016decoding}.
Although the \gls{hicann} architecture has been successfully used to implement deep multi-layer networks using rate-based spiking models\cite{schmitt2016classification} and back-propagation based training, is looses some of its power-efficiency by emulating a Perceptron model. Encoding the activation in the time between spikes can enhance the efficiency significantly\cite{goltz2019fast}.
In all spiking solutions the network operates in continuous time and therefore the size of the network is limited to the number of neurons and synapses available on the chip. 
The \gls{hagen} extension, which is part of the \gls{hicannx} chip, allows a seamless mixture of spiking and non-spiking operation within a single chip.
Since this rate-based operation is based on discrete-time analog vector-matrix mulitplication, a time-multiplexing scheme can be employed, similar to digital accelerators for deep convolutional networks\cite{shawahna2018fpga}. In this case the size of the network is limited only by the size of any external memory.
\begin{figure}
    \center{      \includegraphics[width=0.8\linewidth,page=14,viewport=0 0 205mm 97mm, clip]{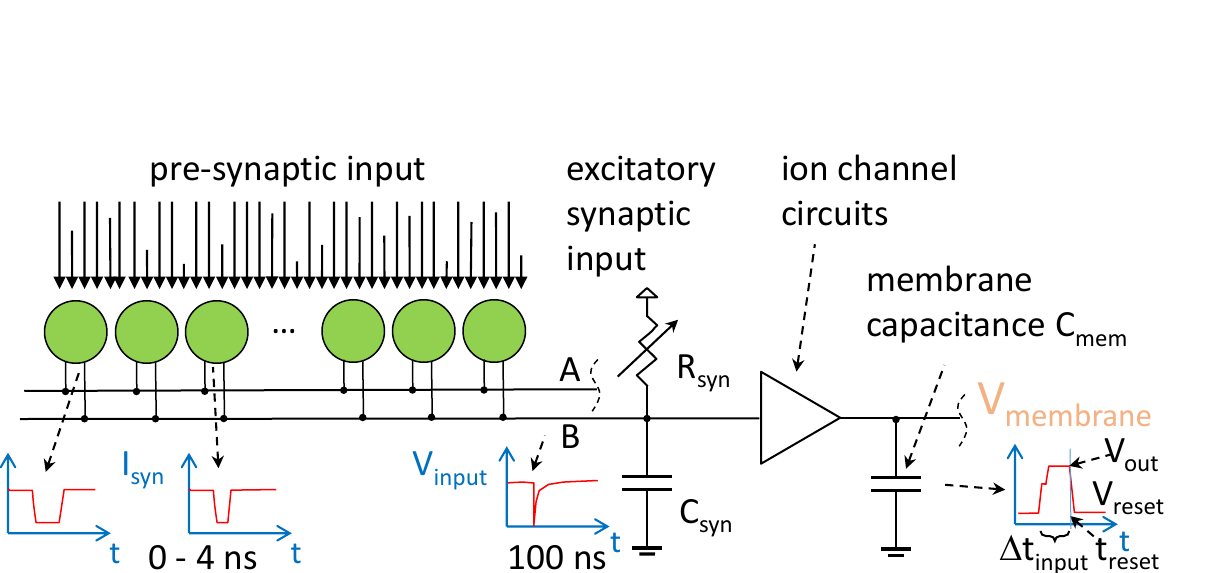}
      \caption{Operating principle of the \gls{hagen} extensions in the \gls{hicannx} chip.
      \label{fig:opneuronx} }}
\end{figure}

Fig.~\ref{fig:opneuronx} visualizes the differences between standard spiking mode and \gls{hagen} mode, which eliminates all temporal dynamics from the neuron. 
By disabling the leakage term of the neuron the membrane just sums up the synaptic input.
The excitatory input is added with a positive and the inhibitory input with a negative sign.
All input is applied during the time interval $\Delta t_\textrm{input}$, after which the membrane voltage is digitized by the \gls{cadc} and the neuron is set to the reset voltage $V_\textrm{reset}$ by a reset signal from the \gls{ppu}.
$\Delta t_\textrm{input}$ can be as short as \SI{100}{ns}.
It depends on the bandwidth of the synaptic input and the number of synaptic rows used, i.e. the total time required to transfer all input events to the synapses. Since the minimum time is at least a few synaptic time constants and nothing is gained by setting the integration time shorter than the conversion time of the \gls{cadc}, a typical value for $\Delta t_\textrm{input}$ is about \SI{500}{ns}.
Thereby, the network can evaluate $2\cdot 10^6\times 256 \times 512 = 2.62\cdot 10^{11}$ multiply-accumulate operations per second. By shortening the conversion time of the \gls{cadc} further speed improvements are possible.
Since the reset voltage of the neuron membrane can be aligned with the lower bound of the \gls{cadc} conversion range the neuron acts like a ReLU unit in this setting\cite{sharma2017era}. 

A standard synapse within BrainScaleS reacts to a pre-synaptic event in a digital fashion: the arrival of a pre-synaptic event generates a fixed current pulse. By enabling short-term facilitation or depression\cite{billaudelle2017stp} the synaptic strength depends on the pre-synaptic firing history. This is achieved by modulating the pulse length generated by the synapse.
Instead of using the firing history, in \gls{hagen} mode the pulse length is transmitted together with the pre-synaptic spike and converted into variable length pulses by the existing \gls{stp} pulse-length modulation circuits.
The digital pulse length information is transmitted by reusing the \SI{5} lower address bits of the \gls{l1} event data, since in the \gls{hagen} mode the network structure is much more regular and not all pre-synaptic address bits are needed.

\begin{figure}
    \center{      \includegraphics[width=1\linewidth,page=26,viewport=0 0 215mm 82mm, clip]{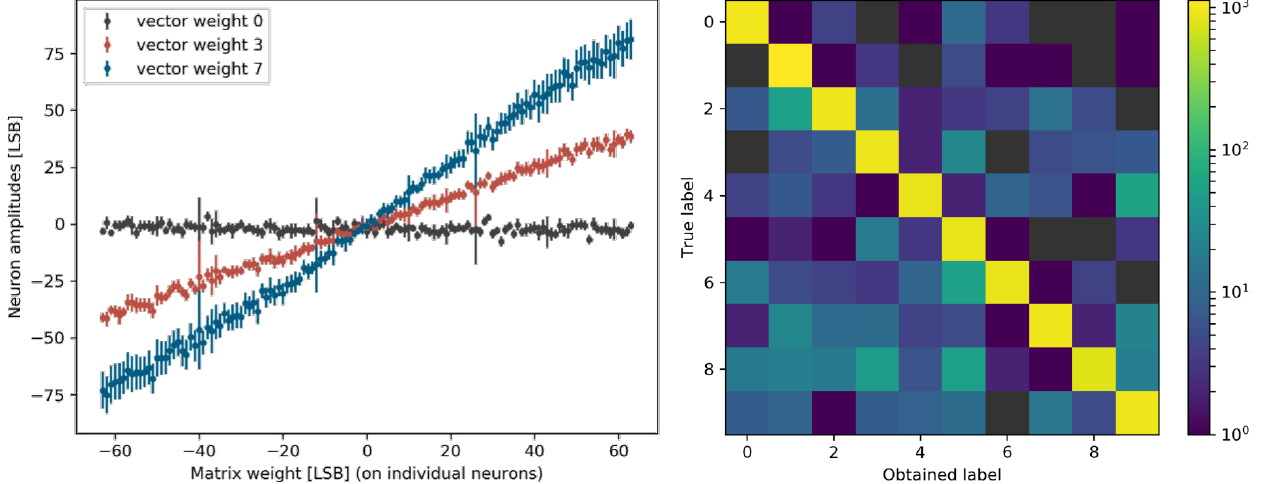}
      \caption{Left: Results for analog vector-matrix multiplication. Right: Confusion matrix for MNIST.
      \label{fig:hagenresults} }}
\end{figure}

Fig.\ref{fig:hagenresults} shows some early results using the activity-based Perceptron mode from \gls{hicannx} for analog vector-matrix multiplication. 
In the left part of the figure, 127 neurons are measured simultaneously. 
Their synaptic weights increase linearly from -63 to 63, i.e. all synapses connected to a single neuron are set to the same weight while the weights incrase from neuron to neuron. 
All synapes receive the same input: 0, 3 or 7 for the black, red and blue traces respectively.
The outputs of all 127 neurons are digitized simultaneously by the \gls{cadc} and the digital values are plotted over the weight values of the neurons.
Although the neuron circuits are calibrated, some fixed-pattern noise remains visible.
The temporal variations are caused by a well-understood circuit flaw, that will be removed in future iterations.

The chip has been subsequently used to perform inference on the MNIST dataset\cite{lecunmnist}. A three-layer network has been trained in Tensorflow\cite{tensorflow2015} to reach a classification rate of 97.43\%.
The weights and input activations of this network have been quantisized to \SI{6}{bit} weight and \SI{5}{bit} input resolution, to fit the trained network to the dynamic range of the analog circuits.
The inference on the test data set has been repeated using the \gls{hicannx} chip.
The resulting classification accuracy was 92.48\%.
The corresponding confusion matrix is shown in the right panel of Fig.~\ref{fig:hagenresults}.
The deterioration is most likely caused by the remaining fixed-pattern noise.
In the future we will include the hardware in the forward-path of the training loop, similar to the approach followed in \cite{schmitt2016classification}, which will most likely improve the accuracy significantly.

\section{Analog Verification of Complex Neuron Circuits\label{sec:teststand}} 
The BrainScaleS systems feature complex mixed-signal circuits to emulate the rich properties of their biological counterparts.
Our neuron circuits, implementing the \gls{adex} equations \cite{brette_05}, possess a multitude of individual subcomponents, such as a leak or adaptation term.
Each of these units is parameterized through a number of digital controls as well as analog voltage and current biases.
Designed to support a variety of different tasks, ranging from biologically realistic firing patterns to analog matrix multiplication, these circuits have to be operated at widely different operating points.
The correct behavior has to be ensured prior to fabrication.
Individual components can often be unit-tested in isolation, making use of convential simulation strategies.
The assesability of a complete design is, however, limited due to error propagation and inter\hyp{}dependencies of parameters.

A suite of benchmark tasks, evaluated on comprehensive testbenches, is required for pre-tapeout verification.
To ensure the required degree of precision over larger arrays of analog circuits, mismatch effects introduced through imperfections in the production process, have to be covered through \gls{mc} simulations.
Different incarnations of a circuit can be obtained by individually fixing the \gls{mc} seed.
These virtual instances can then be characterized, very similar to the fabricated siblings.
Similarly, the worst case behavior can be characterized for the process corners.

In the following paragraphs, we present our simulation strategy and a custom library to aid software-driven simulations within the rich ecosystem of the Python programming language.
We will guide through our benchmarking flow for our current generation of \gls{adex} neurons.
Similar approaches have successfully been taken for the verification of plasticity circuits and vector-matrix muliplication circuits.

\subsection{Interfacing analog simulations from Python}

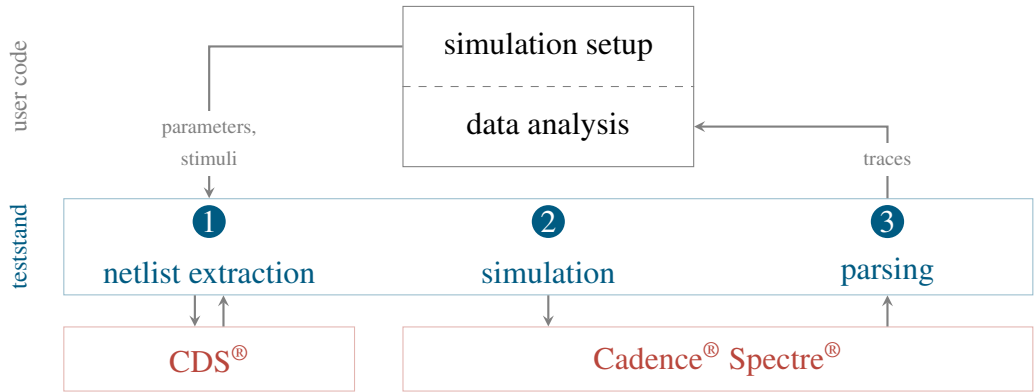
\begin{figure}
	\begin{center}
		\usetikzlibrary{shapes.misc,shapes,fit}
\usetikzlibrary{decorations.pathreplacing}
\usetikzlibrary{arrows.meta,decorations.pathreplacing,fadings,shapes,arrows}

\definecolor{orange}{HTML}{D7AA50}
\definecolor{purple}{HTML}{7A68A6}

\begin{tikzpicture}[x=1cm,y=1cm,scale=0.85,transform shape]
	\definecolor{blue}{RGB}{0,91,130}
	\definecolor{lightblue}{RGB}{110,159,189}
	\definecolor{red}{RGB}{185,70,60}
	\definecolor{lightred}{RGB}{198,141,132}
	\definecolor{green}{RGB}{125,150,110}
	\definecolor{lightgreen}{RGB}{164,181,153}
	\definecolor{orange}{HTML}{D7AA50}
	\definecolor{purple}{HTML}{7A68A6}
	
	\node[gray,rotate=90,scale=1.2] at (-0.7,-1.25) {\footnotesize user code};
	\node[blue,rotate=90,scale=1.2] at (-0.7,-3.75) {\footnotesize teststand};
	
	\node[anchor=north west,minimum width=4.5cm,minimum height=2.5cm,draw=gray] (u1) at (5.25,0) {};
	\node[yshift=+0.625cm,minimum width=4.5cm,align=center] (u11) at (u1)
		{simulation setup};
	\node[yshift=-0.625cm,minimum width=4.5cm,align=center] (u12) at (u1)
		{data analysis};
	\draw[dashed,gray] (u1.west) -- (u1.east);
	
	\node[below right,minimum width=4.5cm,minimum height=1.5cm] (t1) at (0,-3) {};
	\node[shape=circle,inner sep=1pt,fill=blue,yshift=0.8em] at (t1) {\color{white}1};
	\node[yshift=-0.8em] at (t1) {\color{blue}netlist extraction};
	
	\node[below right,minimum width=4.5cm,minimum height=1.5cm] (t2) at (5.25,-3) {};
	\node[shape=circle,inner sep=1pt,fill=blue,yshift=0.8em] at (t2) {\color{white}2};
	\node[yshift=-0.8em] at (t2) {\color{blue}simulation};
	
	\node[below right,minimum width=4.5cm,minimum height=1.5cm] (t3) at (10.5,-3) {};
	\node[shape=circle,inner sep=1pt,fill=blue,yshift=0.8em] at (t3) {\color{white}3};
	\node[yshift=-0.8em] at (t3) {\color{blue}parsing};
	
	\node[below right,minimum width=4.5cm,minimum height=1.0cm,draw=red!40] (c1) at (0,-5) {};
	\node[] at (c1) {\color{red}CDS\textsuperscript{\textregistered}};
	
	\node[below right,minimum width=9.75cm,minimum height=1.0cm,draw=red!40] (c2) at (5.25,-5) {};
	\node[] at (c2) {\color{red}Cadence\textsuperscript{\textregistered} Spectre\textsuperscript{\textregistered}};
	
	\node[below right,minimum width=15cm,minimum height=1.5cm,draw=blue!40] (t) at (0,-3) {};

	\draw[-stealth,thick,gray] (u11.west) -- (u11.west-|t1.north) -- (t1.north);
	\draw[stealth-,thick,gray] (u12.east) -- (u12.east-|t3.north) -- (t3.north);
	
	\draw[-stealth,thick,gray] ([xshift=-1ex]t1.south) -- ([xshift=-1ex]c1.north);
	\draw[-stealth,thick,gray] ([xshift=+1ex]c1.north) -- ([xshift=+1ex]t1.south);
	
	\draw[-stealth,thick,gray] (t2.south) -- (t2.south |- c2.north);
	\draw[stealth-,thick,gray] (t3.south) -- (t3.south |- c2.north);

	\node[anchor=south,fill=white,align=center] at ([yshift=+0.7em]t1.north) {\footnotesize\color{gray}parameters,\\[-0.2em]\footnotesize\color{gray}stimuli};
	\node[anchor=south,fill=white,align=center] at ([yshift=+0.7em]t3.north) {\footnotesize\color{gray}traces};
\end{tikzpicture}

	\end{center}
	\caption{Structure of a teststand-based simulation highlighting the interaction with the Cadence Design Suite. Image taken from \cite{gruebill2020bss2methods}.}
	\label{fig:schematic_teststand}
\end{figure}

\begin{figure*}
	\begin{center}
		\includegraphics[width=\textwidth]{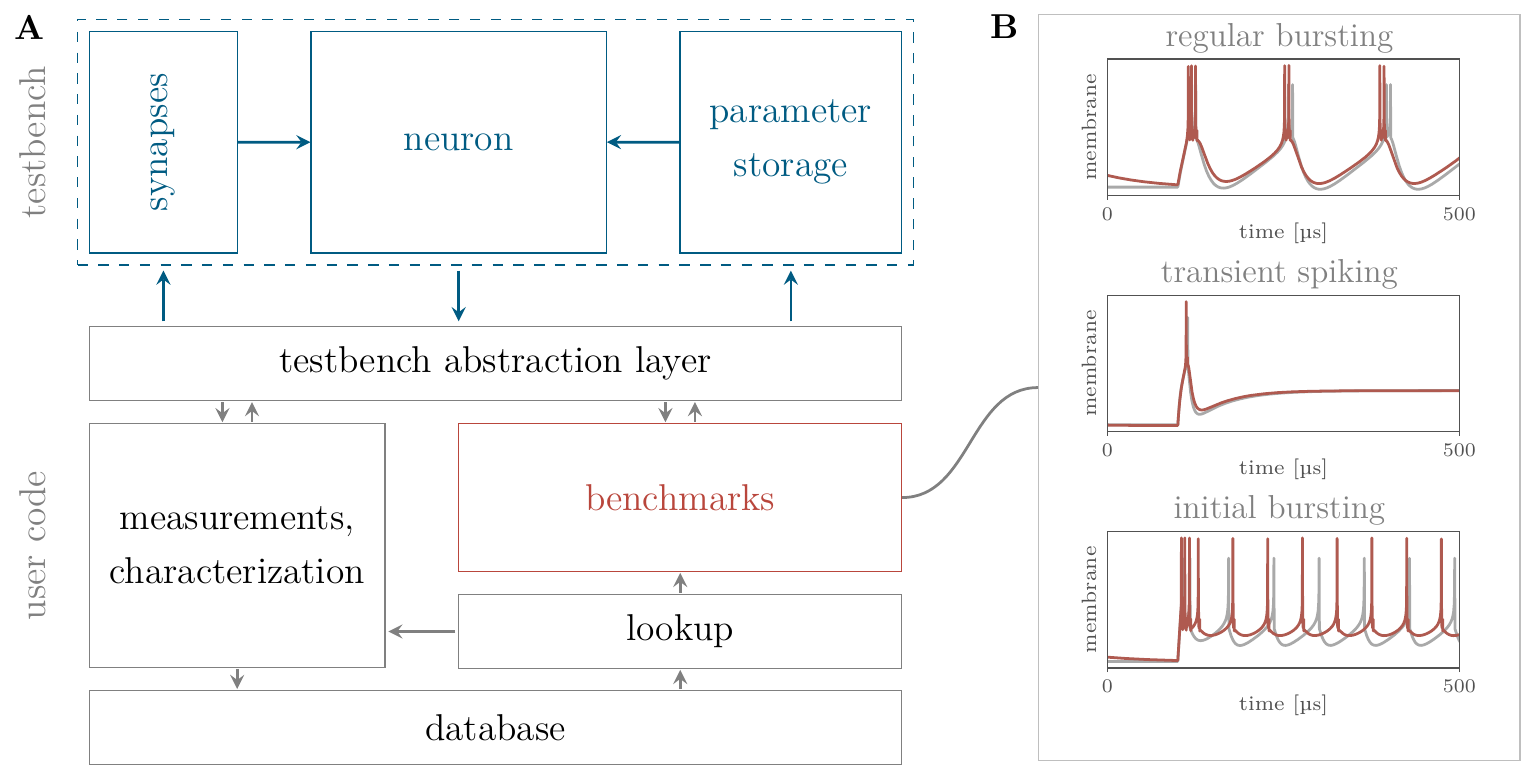}
	\end{center}
	\caption{\gls{mc} calibration workflow of an \gls{adex} neuron circuit using teststand. \textbf{A}~Testbench and overview on the software stack for characterization and calibration. \textbf{B}~Membrane traces (red) of a neuron circuit simulation configured for regular bursting, transient spiking, and initial bursting. The results of a numerical integration of the \gls{adex} equations is shown as a reference (gray).}
	\label{fig:teststand_calibration}
\end{figure*}

Our custom Python module \emph{teststand} provides a tight integration between analog circuit simulations and the ecosystem of the programming language\cite{gruebill2020bss2methods}.
It mainly consists of a software layer to interface with the Cadence Spectre simulator and other tools from the Cadence Design Suite.

Teststand extracts the testbench's netlist directly from the target cell view as available in the design library.
The data is accessed by querying the database via an OCEAN script executed as a child process.
Teststand then reads the netlist and modifies it according to the user's specification.
In addition to the schematic description, Spectre netlists also contain simulator instructions.
Teststand generates these statements according to the user's Python code.
Specifically, the user can define analyses to be performed by the simulator, such as DC, AC, and transient simulations.
\gls{mc} analyses are supported as well and play an important role in the verification strategies presented below.
Teststand can be easily extended to support all features provided by the backend.

All circuit parameters, stimuli, and nodes to be recorded are specified using an object-oriented interface that resembles Spectre simulation instructions.
\begin{lstlisting}[style=pythonstyle]
cell = ('mylib', 'mycell', 'schematic')
nets = ['I0.mynet']

teststand = Teststand(cds_lib, cell)
tran = TransientAnalysis('tran', 1e-3)
simulation = Simulation(
	[tran], params, save=nets)
result = teststand.simulate(simulation)
\end{lstlisting}
The \texttt{simulate()}-call executes Spectre as a child process.
Basic parallelization features are natively provided via the \textit{multiprocessing} library.
Scheduling can be trivially extended to support custom compute environments.
The simulation log is parsed and potential error messages are presented to the user as Python exceptions.

Results are read and provided to the user as structured NumPy arrays.
This allows to resort to the vast amount of data processing libraries available in the Python ecosystem to process and evaluate recorded data.
Most notably, this includes NumPy \cite{oliphant2006guide}, SciPy \cite{Scipy2001}, and Matplotlib \cite{hunter2007matplotlib}.
As a side effect, the latter allows to directly generate rich publication-ready figures from analog circuit simulations.

\subsection{Monte Carlo calibration of \gls{adex} neuron circuits}

As shown in Fig.~\ref{fig:teststand_calibration}A, we used the teststand library, inter alia, for the verification of our \gls{adex} design.
The model equations feature a high-dimensional parameter space, allowing for a wide range of behaviors.
Our circuit, on the other hand, is parameterized through 24 individual analog bias sources and a set of digital controls.
Starting from first-order models of the utilized subcomponents, we characterized the circuit's dynamics through a set of measurements on the full neuron circuit.
With the results stored in a database, we established a transformation between between the circuits's and the models's parameter spaces.
The influence of mismatch effects manifests itself in deviations in these calibration curves for individual neuron instances.
We applied the above framework for a large number of neuron incarnations, obtained by fixing the respective \gls{mc} seeds.

The circuit was benchmarked against multiple firing patterns, such as \emph{transient spiking}, \emph{regular bursting}, and \emph{initial bursting} \cite{naud08}.
For each of these targets, a set of biases, corresponding to the respective parameter set from literature, was determined through a reverse lookup based on the above transformations.
Examplary results for a single neuron simulation are shown in Fig.~\ref{fig:teststand_calibration}~B.

The presented approach enforces the development of calibration algorithms before tape-out.
Especially for circuits with large parameter spaces, there might occur multi-dimensional dependencies which can be hard to resolve.
The strategy might also reveal an insufficient parametrization not necessarily apparent from individual unit tests.
In order to uncover potential regressions due to modifications to a circuit, simulations based on teststand can easily be automated and allow continuous integration testing for full-custom designs.

\section{Conclusion\label{sec:conclusion}}

The development and implementation of the presented second generation BrainScaleS architecture will hopefully continue during the next years.
The outcome we hope for is a multi-wafer system, constructed from hundreds of \SI{30}{cm} silicon wafers, each one directly embedded in a \gls{pcb} and all of them interconnected to form a novel large-scale analog neuromorphic platform.
A system capable of answering questions about learning and development in large scale, biologically realistic neural networks.

Utilitzing standard \gls{cmos} technology to build large-scale analog accelerated neuromorphic hardware systems places our approach in the middle between the two major research directions for AI circuits: digital accelerators and novel persistent memory devices. 
It presents a complementary option to theses technologies. Compared to systems based on novel device technology it has advantages, like the high operational speed, low energy requirements for learning, the possibility to use any standard \gls{cmos} process without regards to back-of-the-line compatibility and the capability to replicate relevant biological structures more easily.
In comparison to digital implementations, like Loihi or SpiNNaker \cite{davies2018loihi, furber2014spinnaker}, the fully analog implementation of complex neural structures combined with true in-memory computing allows for time-continuous emulation of neural dynamics and much higher emulation speed at similar energy efficiences.
Most importantly, analog \gls{cmos} implementations might be the essential step to uncover the learning rules needed to cope with substrate variations. 
In our systems the local learning rules do not only train the system to perform a certain task, but simultaneously adjust the operating point of the circuits and compensate fixed pattern noise\cite{billaudelle2017stp}. 
This will be an essential property for future novel computing systems based on advanced device technologies as well, since they all are expected to have substantially increased device-to-device variations.
We hope that our \gls{bss} platform will help to gain insight in the necessary algorithms in the upcoming future.

In the short term the \gls{bss} system allows the combination of energy- and cost-efficient analog inference with local learning rules for a multitude of practical applications, scaling from small systems for edge computing up to high-performance neuromorphic cloud computing.

\section{Acknowledgments}
The authors wish to express their gratitude to Andreas Gr\"ubl, Yannik Stradmann, Vitali Karasenko, Korbinian Schreiber, Christian Pehle, Ralf Achenbach, Markus Dorn and Aron Leibfried for their invaluable help and active contributions in the development of the BrainScaleS~2 \glspl{asic} and systems.\\
They are not forgetting the important role their former colleagues Andreas Hartel, Syed Aamir, Gerd Kiene, Matthias Hock, Simon Friedmann, Paul M\"uller, Laura Kriener and Timo Wunderlich had in these endeavors.\\
They also want to thank their collaborators Sebastian H\"oppner from TU Dresden and Tugba Demirci from EPFL Lausanne for their contributions to the BrainScaleS~2 prototype \gls{asic}.\\
Very special thanks go to Eric M\"uller and his team for leading the software development as well as Mihai Petrovici, Sebastian Schmitt and the late Karlheinz Meier for their invaluable advice.\\
This work has received funding from the European Union
Seventh Framework Programme ([FP7/2007-2013]) under
grant agreement no 604102 (HBP rampup), 269921 (BrainScaleS),
243914 (Brain-i-Nets), the Horizon 2020 Framework Programme
([H2020/2014-2020]) under grant agreement 720270 and 785907 (HBP SGA1 and SGA2)
as well as from the Manfred St\"ark Foundation.\\
\section{Author Contribution}
J.S. created the concept, has been the lead architect of the \gls{bss} systems and wrote the manuscript except for Section~\ref{sec:teststand}, which was written by S.B.
S.B. also created the teststand software and conceived the simulations jointly with P.D, who performed the simulations and prepared the results. J.W. performed the measurements for the \gls{hagen} mode and created figure~\ref{fig:hagenresults}. All authors edited the manuscript together.

\end{document}